\pdfoutput=1

\documentclass[11pt]{article}
\usepackage[]{emnlp2021}

\usepackage{times}
\usepackage{latexsym}
\usepackage{amssymb}
\usepackage{graphicx}
\usepackage{algorithm}
\usepackage{algorithmicx}
\usepackage{algpseudocode}
\usepackage{amsmath}
\usepackage{booktabs}
\usepackage{multirow}
\usepackage{comment}
\usepackage{epstopdf}
\usepackage{makecell}
\usepackage[T1]{fontenc}

\usepackage[utf8]{inputenc}

\usepackage{microtype}

\title{Weakly-supervised Text Classification Based on Keyword Graph}

\author{
	Lu Zhang$^{1}$\thanks{~~The author did most work during internship at  Alibaba.} \quad Jiandong Ding$^2$ \quad  Yi Xu$^{1}$\footnotemark[1] \quad Yingyao Liu$^2$ \quad Shuigeng Zhou$^1$\thanks{~~Correspondence author} \\
	$^1$Shanghai Key Lab of Intelligent Information Processing, and School of \\ Computer Science, Fudan University, China\\
	$^2$Alibaba Group \\
	{\tt\small \{l\_zhang19, jdding, yxu17, sgzhou\}@fudan.edu.cn
	}
}

\begin{document}
\maketitle
\begin{abstract}
Weakly-supervised text classification has received much attention in recent years for it can alleviate the heavy burden of annotating massive data.
Among them, keyword-driven methods are the mainstream where user-provided keywords are exploited to generate pseudo-labels for unlabeled texts.
However, existing methods treat keywords independently, thus ignore the correlation among them, which should be useful if properly exploited.
In this paper, we propose a novel framework called \emph{ClassKG} to explore keyword-keyword correlation on keyword graph by GNN.
Our framework is an iterative process. In each iteration, we first construct a keyword graph,
so the task of assigning pseudo labels is transformed to annotating keyword subgraphs.
To improve the annotation quality, we introduce a self-supervised task to pretrain a subgraph annotator, and then finetune it.
With the pseudo labels generated by the subgraph annotator, we then train a text classifier to classify the unlabeled texts.
Finally, we re-extract keywords from the classified texts.
Extensive experiments on both long-text and short-text datasets show that our method substantially outperforms the existing ones.



\end{abstract}

\section{Introduction}
\emph{Text classification} is one of the most fundamental tasks in natural language processing (NLP).
In real-world scenarios, labeling massive texts is time-consuming and expensive, especially in some specific areas that need domain experts to participate.
\emph{Weakly-supervised text classification} (WTC) has received much attention in recent years because it can substantially reduce the workload of annotating massive data.
Among the existing methods, the mainstream form is keyword-driven~\cite{keywords_2000,keywords_2003,keywords_2006,WeSTClass,AAAI19_hierarchical,LOTClass,Conwea,X-class,taxoclass}, where the users need only to provide some keywords for each class. Such class-relevant keywords are then used to generate pseudo-labels for unlabeled texts.

\begin{figure}[t]
	\centering
	\includegraphics[width=1.0\columnwidth]{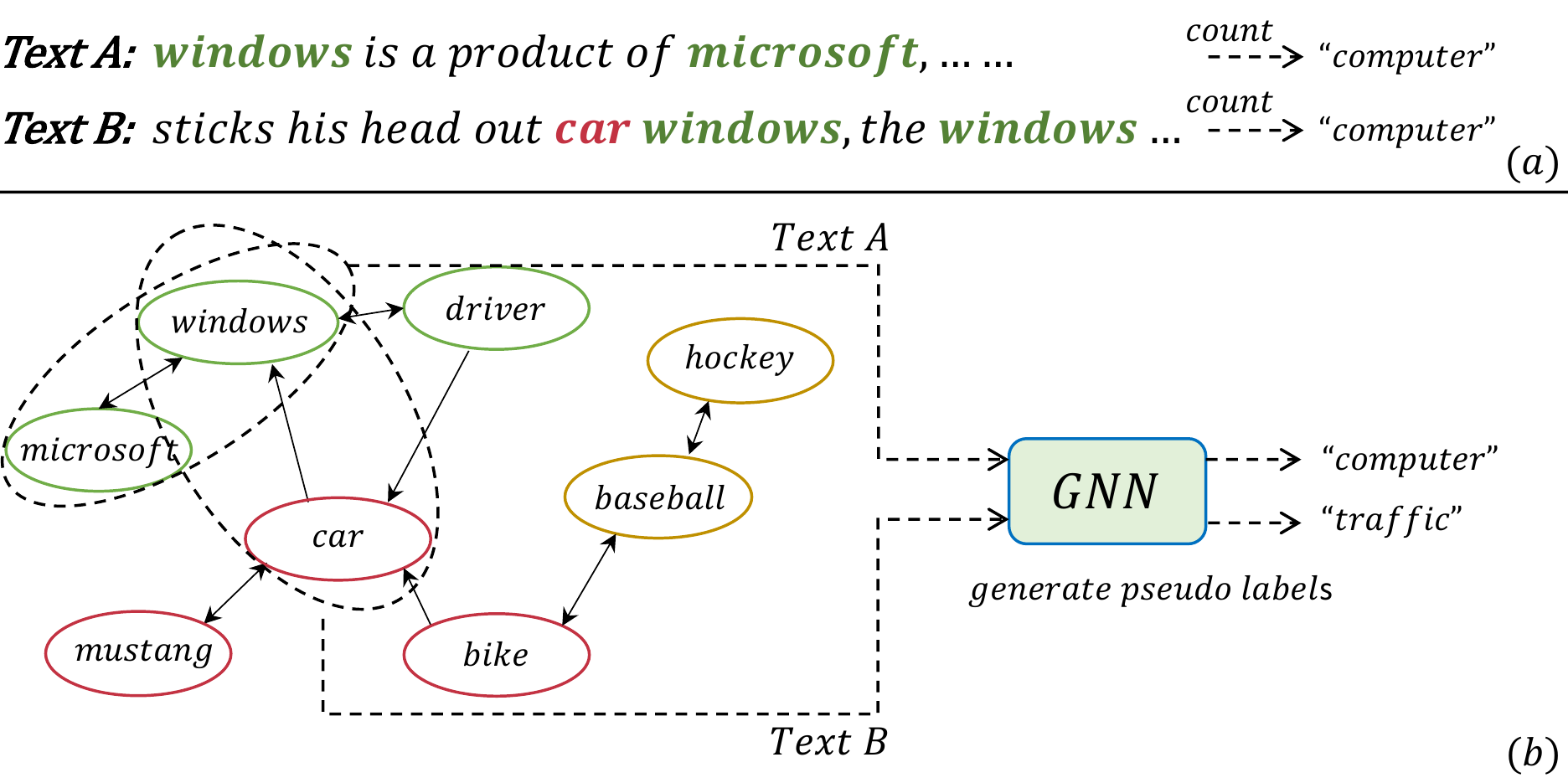} %
	\caption{(a) Existing methods do not consider the correlation among keywords, which will generate wrong pseudo-label for text B.
		(b) Our method exploits the correlation among keywords by GNN over a keyword-graph, and converts the task of assigning pseudo-labels for unlabeled texts to annotating subgraphs, which leads to much better performance.}
	\label{fig_introduction}
\end{figure}

Keyword-driven methods usually follow an iterative process:
generating pseudo-labels using keywords, building a text classifier, and updating the keywords or self-training the classifier.
Among them, the most critical step is generating pseudo-labels.
Most existing methods generate pseudo-labels by counting keywords, with which
the pseudo-label of a text is determined by the category having the most keywords in the text.

However, one major drawback of these existing methods is that \emph{they treat keywords independently, thus ignore their correlation}. Actually, such correlation is important for the WTC task if properly exploited, as a keyword may implies different categories when it co-occurs in texts with other different keywords.
As shown in Fig.~\ref{fig_introduction}, suppose the users provide keywords “windows” and “microsoft” for class ``\textit{computer}'' and ``car” for class ``\textit{traffic}''.
When “windows” and “microsoft” appear in text A, the ``windows'' means operating system, and text A should be given a pseudo-label of ``\textit{computer}''.
However, when “windows” meets ``car” in text B, the ``windows” means the windows of a car and text B should be given a pseudo-label ``\textit{traffic}''.
With previous simple keyword counting, text A can get a correct pseudo-label, but text B cannot.
Therefore, treating keywords independently is problematic.

In this paper, we solve the above problem with a novel iterative framework called \emph{\textbf{ClassKG}} (the abbreviation of \emph{\textbf{Class}}ification with \emph{\textbf{K}}eyword \emph{\textbf{G}}raph) where the keyword-keyword relationships are exploited by GNN on \textbf{k}eyword \textbf{g}raph. In our framework, the task of assigning pseudo-labels to texts using keywords is transformed into annotating keyword subgraphs.
Specifically, we first construct a keyword graph $\mathcal{G}$ with all provided keywords as nodes and each keyword node updates itself via its neighbors.
With $\mathcal{G}$, any unlabeled text $T$ corresponds to a subgraph $\mathcal{G_T}$ of $\mathcal{G}$, and assigning a pseudo-label to $T$ is converted to annotating subgraph $\mathcal{G_T}$.
To accurately annotate subgraphs, we adopt a paradigm of first self-supervised training and then finetuning. The keyword information is propagated and incorporated contextually during keyword interaction.
We design a self-supervised pretext task that is relevant to the downstream task, with which the finetuning procedure is able to generate more accurate pseudo-labels for unlabeled texts.
Texts that contains no keywords are ignored.
With the pseudo-labels, we train a text classifier to classify all the unlabeled texts.
And based on the classification results, we re-extract the keywords, which are used in the next iteration.

Furthermore, we notice that some existing methods employ simple TF-IDF alike schemes for re-extracting keywords, which makes the extracted keywords have low coverage and discrimination over the unlabeled texts. 
Therefore, we develop an improved keyword extraction algorithm that can extract more discriminative keywords to cover more unlabeled texts, with which more accurate pseudo-labels can be inferred.

In summary, our contributions are as follows:
\begin{itemize}
\item
We propose a new framework \emph{ClassKG} for weakly supervised text classification where the correlation among different keywords is exploited via GNN over a keyword graph, and the task of assigning pseudo-labels for unlabeled texts is transformed into annotating keyword subgraphs on the keyword graph.
\item
We design a self-supervised training task on the keyword graph, which is relevant to the downstream task and thus can effectively improve the accuracy of subgraph annotating.
%
%
\item
We conduct extensive experiments on both long text and short text benchmarks. Results show that our method substantially outperforms the existing ones.
\end{itemize}

\section{Related Work}
Here we review the related works, including weakly-supervised text classification and self-supervised learning.

\subsection{Weakly-Supervised Text Classification}

Weakly-supervised text classification (WTC) aims to use various weakly supervised signals to do text classification. Weak supervision signals used by existing methods includes external knowledge base~\cite{distant_supervision,Dataless,dataless_hierarchical,Zero-shot_Text_Classification}, keywords~\cite{keywords_2000,keywords_2003,keywords_2006,Doc2Cube,WeSTClass,AAAI19_hierarchical,LOTClass,Conwea,X-class,taxoclass} and heuristics rules~\cite{DP,Snorkel,DP_ACL_Discourse,email_intent}.
In this paper, we focus on keyword-driven methods.
Among them, WeSTClass~\cite{WeSTClass} introduces a self-training module that bootstraps on real unlabeled data for model refining. WeSHClass~\cite{AAAI19_hierarchical} extends WeSTClass to hierarchical labels. LOTClass~\cite{LOTClass} uses only label names as the keywords. Conwea~\cite{Conwea} leverages contextualized corpus to disambiguate keywords.
However, all these methods treat keywords independently, so ignore their correlation, which is actually useful information for generating pseudo-labels.
Different from the existing methods, we exploit keyword correlation by applying GNN to a keyword graph, which can significantly boost the quality of pseudo-labels.

\subsection{Self-supervised Learning}
Self-supervised learning exploits internal structures of data and formulates predictive tasks to learn good data representations.
The key idea is to define a pretext task and generate surrogate training samples automatically to train a model.
A wide range of pretext tasks have been proposed.
For images, self-supervised strategies include predicting missing parts of an image~\cite{context_encoders}, patch orderings~\cite{SSL_jigsaw_puzzles} and instance discrimination~\cite{Moco}.
For texts, the tasks can be masked language modeling~\cite{BERT}, sentence order prediction~\cite{ALBERT} and sentence permutation~\cite{BART}.
For graphs, the pretext tasks can be contextual property prediction~\cite{Graph_Transformer}, attribute and edge generation~\cite{Gpt-gnn}.

However, up to now there are only a few works of self-supervised learning for subgraph representation~\cite{Sub_Graph_Contrast,GCC}, where contrastive loss is used for subgraph instance discrimination, and the downstream tasks they serve are mainly
whole graph classification or node classification, instead of subgraph classification.
Here, we design a new pretext task based on the keyword graph to learn better representations of keyword subgraphs, with which the accuracy of pseudo-label generation is improved, and consequently classification performance is boosted.

\begin{figure*}[t]
	\centering
	\includegraphics[width=1.0\textwidth]{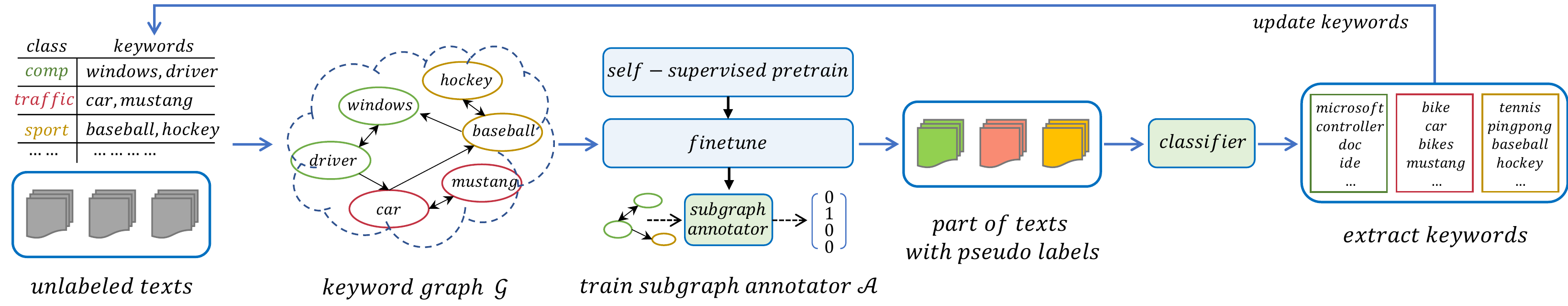} %
	\caption{Our framework follows an iterative paradigm. In each iteration, we first build a keyword graph $\mathcal{G}$, with which unlabeled texts corresponds to subgraphs of $\mathcal{G}$, and assigning pseudo-labels to texts is transformed to annotating the corresponding subgraphs. To train the subgraph annotator, we design a self-supervised pretext task, and finetune it. Then, a text classifier is trained with the pseudo-labels. Based on the classification results of the classifier, keywords are re-extracted and updated for the next iteration.}
	\label{framework}
\end{figure*}

\section{Method}

\subsection{Problem Definition}
The input data contains two parts:
a) A set of user-provided initial keywords $S =\{S_1,S_2,...S_C\}$ for $C$ categories, where $S_i = \{w^i_{1},w^i_{2},...,w^i_{k_i}\}$ denotes the $k_i$ keywords of category $i$,
b) A set of $n$ unlabeled texts $U=\{U_1,U_2,...,U_n\}$ falling in $C$ classes.
Our aim is to build a text classifier and assign labels to the unlabeled texts $U$.

\subsection{Framework}
Fig.~\ref{framework} is the framework of our method, which follows an iterative paradigm.
In each iteration,
we first build a keyword graph $\mathcal{G}$ based on the co-occurrence relationships between keywords from all classes.
Each keyword node aggregates information from its neighbors.
Here, an unlabeled text corresponds to a subgraph of $\mathcal{G}$ and annotating unlabeled texts is converted to annotating the corresponding subgraphs.
To train a high-quality subgraph annotator $\mathcal{A}$,
we first train $\mathcal{A}$ with a designed self-supervised task on $\mathcal{G}$, then finetune it with noisy labels.
After that, unlabeled texts containing keywords are mapped to subgraphs, which are annotated by $\mathcal{A}$ to generate pseudo-labels.
With the pseudo-labels, we train a text classifier to classify all unlabeled texts. Based on the classification results, keywords are re-extracted and updated for the next iteration, until the keywords change little. 

\subsection{Keyword Graph Construction}
To model the relationships among keywords, we construct a keyword graph $\mathcal{G}$ by representing keywords as vertices and co-occurrences between keywords as edges, denoted as $\mathcal{G} = (\mathcal{V},\mathcal{E})$.

For vertices $\mathcal{V}$, the embedding of a node is initialized with vector $x_v=[v_{class};v_{index}] \in \mathbb{R}^{C + |\mathcal{V}|}$, where $v_{class}$ is the one-hot embedding of keyword class, $v_{index}$ is the one-hot embedding of keyword index, and $C$ is the number of classes.

For edges $\mathcal{E}$, if keywords $w_i$ and $w_j$ occur in an unlabeled text in order, there exists a directed edge from $w_i$ to $w_j$. Meanwhile, we take their co-occurrences $F_{ij}$ in unlabeled texts as edge attribute.
Considering the limited number of keywords contained in a text, we do not use any sliding window to limit the number of edges.

With the keyword graph $\mathcal{G}$, keyword feature information is propagated and aggregated by GNN.

\subsection{Subgraph Annotator Training}
With the keyword graph, an unlabeled text is converted into a subgraph of $\mathcal{G}$.
Specifically, the keywords in a text hit a set of vertices in $\mathcal{G}$. The subgraph is the induced subgraph of the hit vertices in $\mathcal{G}$.
Assigning pseudo labels to unlabeled texts is equivalent to annotating the corresponding subgraphs, which is a graph-level classification problem.
In practice, we employ graph isomorphism network (GIN)~\cite{GIN} as our subgraph annotator to perform node feature propagation and subgraph readout.
The keyword feature is propagated and aggregated as follows:
\begin{equation}\small
	h_v^{(k)} = MLP^{(k)}\big( (1 + \varepsilon^{(k)} ) \cdot h_v^{(k-1)}  +  \sum\nolimits_{u \in N(v)} {h_u^{(k-1)}}   \big)
\end{equation}
where $h_v^{(k)}$ denotes the representation of node $v$ after the $k^{th}$ update. $MLP^{(k)}$ is a multi-layer perceptron in the $k^{th}$ layer. $\varepsilon$ is a learnable parameter. $N(v)$ denotes all the neighbors of node $v$.
Then, we perform readout to obtain subgraph representation:
\begin{equation}\small
	h_{G} = CONCAT\big(SUM(\{ h_v^{(k)}|v \in G \}) | k = 0...K   \big)
\end{equation}
GIN concatenates the sum of all node features from the same layer as the subgraph representation.

To train a subgraph annotator with high annotating accuracy, we first train a GIN via a designed self-supervised task, then finetune it.

\subsubsection{Self-supervised Training on Graph}
As mentioned in previous self-supervised learning works~\cite{GNN_Pretrain_Strategies}, a successful pre-training needs examples and target labels that are correlated with the downstream task of interest. Otherwise, it may harm generalization, which is known as \textit{negative transfer}.
Considering that the downstream task is a graph-level classification, we design a graph-level self-supervised task, which is highly relevant to subgraph annotation.
Our self-supervised method is shown in Alg.~\ref{alg:ssl},
where the subgraph annotator $\mathcal{A}$ learns to predict the class of the start point of a random walk and the subgraph derived from the random walk is similar to the subgraph generated by an unlabeled text.

To begin with, we randomly sample a keyword $w_r$ from class $C_r$ as the start point of a random walk.
The number of random walk steps follows the same Gaussian distribution $\mathcal{N}(u_s,\sigma^2_s)$ as that of the number of keywords appearing in an unlabeled text in $U$.
Therefore, we estimate the parameters of the Gaussian distribution $u_s,\sigma^2_s$ based on $U$ as follows:
\begin{align}
	\label{Gaussian_distribution}
	u_s &= \frac{1}{n}\sum\limits_i^n {kf({U_i})}  \\
	\sigma^2_s &= \frac{1}{{n - 1}}\sum\limits_i^n {{{[kf({U_i}) - {u_s}]}^2}}
\end{align}
where $kf({U_i})$ is the number of keywords contained in text $U_i$.
Then, we can sample the length $L$ of random walk from distribution $\mathcal{N}(u_s,\sigma^2_s)$.
%
The probability of walking from node $w_i$ to node $w_j$ is derived from the co-occurrence frequency by
\begin{equation}
	{p_{ij}} = \frac{{{F_{ij}}}}{{\sum\nolimits_{{w_k} \in N({w_i})} {{F_{ik}}} }}
\end{equation}
where $F_{ik}$ is the co-occurrence frequency of $w_i$ followed by $w_k$. $N(w_i)$ is the neighbors set of $w_i$.

Then, we start from node $w_r$ to perform a $L$-step random walk. In each step, $p_{ij}$ determines the probability of jumping from $w_i$ to neighbor $w_j$.
At the end of random walk, we obtain a subgraph $G_r$, which is the induced subgraph of the traversed nodes in the keyword graph $\mathcal{G}$.

Our self-supervised task is designed to take $G_r$ as the input of $\mathcal{A}$ and make  $\mathcal{A}$ learn to predict the class of start point $w_r$. The loss function is defined as the negative log likelihood of ${C_r}$:

\begin{equation}
	\mathcal{L}_{SSL} = - \sum\limits_{r \in rand} {{C_r} \log (\mathcal{A}(G_r)) }
\end{equation}


\newcommand{\algrule}[1][.4pt]{\par\vskip.1\baselineskip\hrule height #1\par\vskip.3\baselineskip}
\begin{algorithm}[t]
	\caption{Self-supervision on Keyword Graph}
	\hspace*{0.02in} {\bf Input:}
	keyword graph $\mathcal{G}$, unlabeled texts $U$, Gaussian parameters $u_s,\sigma^2_s$, edge probability $p_{ij}$\\
	\hspace*{0.02in} {\bf Output:}
	pretrained subgraph annotator $\mathcal{A}$
	\begin{algorithmic}[1]
		\algrule
		\Repeat
		\State Randomly sample a class $C_{r}$.
		\State Sample a keyword $w_{r}$ from class $C_{r}$.
		\State Sample $L$ from  distribution $\mathcal{N}(u_s,\sigma^2_s)$.
		\State Perform a random walk on $\mathcal{G}$, with $w_{r}$ as start point, $p_{ij}$ as probability, $L$ as length,
		which will obtain a subgraph $G_r$.
		\State With $G_r$ as input of $\mathcal{A}$, $C_{r}$ as predicted target, compute the loss.
		\State Compute the gradient and update parameters of  $\mathcal{A}$.
		\Until{convergence}

	\end{algorithmic}
	\label{alg:ssl}
\end{algorithm}


\subsubsection{Finetuning}
After pre-training the subgraph annotator $\mathcal{A}$ , we finetune it for a few epochs.
The labels of finetuning are generated by voting as follows:
\begin{equation}
	{\hat y_i} = \arg \mathop {\max }\limits_k \{ \sum\limits_j {tf({w_j},{U_i})} |\forall ({w_j} \in {S_k})\}
\end{equation}
where $tf({w_j},{U_i})$ denotes the term-frequency (TF) of keyword $w_j$ in text $U_i$.
The loss function is defined as follows:
\begin{equation}
	\mathcal{L}_{FT} = - \sum\limits_{i=1}^n {(kf(U_i) > 0) {\hat y_i} \log (\mathcal{A}(G_i)) }
\end{equation}
where 
$G_i$ is the subgraph of text $U_i$.
Note that the number of epochs for fine-tuning cannot be too large, otherwise it may degenerate into voting.

\subsection{Text Classifier}
After training the subgraph annotator $\mathcal{A}$, we use it to annotate all the unlabeled texts $U$ and generate pseudo-labels, which are used to train a text classifier.
Texts containing no keywords are ignored.
Our framework is compatible with any text classifier.
We use Longformer~\cite{Longformer} as the long text (document) classifier and BERT~\cite{BERT} as the short text (sentence) classifier.
Following previous works~\cite{WeSTClass,LOTClass}, we self-train~\cite{self-training} the classifier on all unlabeled
texts.
The predicted labels for all unlabeled texts by the text classifier are then used to re-extract keywords.

\subsection{Keywords Extraction}
Considering that the coverage and accuracy of user-provided keywords are limited, we re-extract keywords based on the predictions of the text classifier in each iteration.
Existing methods use indicators such as term frequency (TF), inverse document frequency (IDF) and their combinations~\cite{Conwea} to rank words, and a few top ones are taken as keywords.
However, they treat all indicators equally, and are prone to select common and low-information words.
Here, we employ an improved TF-IDF scheme, which increases the significance of IDF to reduce the scores of common words.
The score of word $w_i$ in class $C_k$ is evaluated as follows:
\begin{equation}
	Q(w_i,C_k) = TF(w_i,C_k) \cdot IDF(w_i)^M
	\label{eq:word_score}
\end{equation}
Above, $M$ is a hyperparameter.
According to the score, we select the top $Z$ words in each category as the keywords for the next iteration.



To determine whether the model has converged,
we define the change of keywords as follows:
\begin{equation}
	\Delta = \frac{{|{S^{{T_i}}} - {S^{{T_i}}} \cap {S^{{T_{i - 1}}}}|}}{{|{S^{{T_i}}}|}}
\end{equation}
where $S^{{T_i}}$ is the keywords set of the $i^{th}$ iteration.
If $\Delta < \epsilon$ (a hyperparameter), the iteration stops.

\section{Experiments}
\subsection{Datasets}
Different from existing works that do experiments only on long texts or short texts, we verify our method on both long and short texts.
For long texts, we use two news datasets: \textit{The 20 Newsgroups} and \textit{The New York Times}, both of which are news articles and can be partitioned into fine-grained classes and coarse-grained classes.
For short texts, we use four benchmark datasets: \textit{AG News}~\cite{AG_News}, \textit{DBPedia}~\cite{DBPedia}, \textit{IMDB}~\cite{IMDB} and \textit{Amazon}~\cite{Amazon}.
Tab.~\ref{tab:Datasets_statistics} gives the statistics of the datasets.
The initial keywords and evaluation settings follow previous works.
For long texts, the initial keywords follow Conwea~\cite{Conwea} and evaluation results on the entire datasets are reported.
For short texts, we follow LOTClass~\cite{LOTClass} and use the label names as initial keywords. The evaluation is performed on the test set.
For all classes, we use no more than four keywords per category.

\def\hs2{0.3em}
\begin{table}[t]
	\centering 	
	\scalebox{0.9}{
		\begin{tabular}{c@{\hspace{\hs2}}|c@{\hspace{\hs2}}|c@{\hspace{\hs2}}|c@{\hspace{\hs2}}}
			\toprule
			Datasets  & \# Texts &  \# Classes & Avg Len \\
			\midrule
			20News          & 18,846 & 20 (F) or 7 (C) &  400   \\
			NYT             & 13,081 & 26 (F) or 5 (C)  &  778   \\
			AG News         & 127,600&  4   &   54     \\
			DBPedia         & 630,000&  14  &   70 \\
			IMDB            & 50,000 &  2   &  297 \\
			Amazon          &4,000,000& 2   &  104   \\
			\bottomrule
		\end{tabular}
	}
	\caption{Dataset statistics. `F': Fine. `C': Coarse
	}
	\label{tab:Datasets_statistics}
\end{table}

\def\hs1{0.5em}
\begin{table*}[ht]
	\centering 	
	\scalebox{0.93}{
		\begin{tabular}{c@{\hspace{\hs1}}|c@{\hspace{\hs1}}c@{\hspace{\hs1}}c@{\hspace{\hs1}}c@{\hspace{\hs1}}|c@{\hspace{\hs1}}c@{\hspace{\hs1}}c@{\hspace{\hs1}}c@{\hspace{\hs1}}}
			\toprule
			& \multicolumn{4}{c}{20 Newsgroup}
			& \multicolumn{4}{|c}{NYT}
			\\
			\hline
			\multirow{2}{*}{Methods}
			& \multicolumn{2}{c}{Fine-grained}
			& \multicolumn{2}{c|}{Coarse-grained}
			& \multicolumn{2}{c}{Fine-grained}
			& \multicolumn{2}{c}{Coarse-grained}
			
			\\
			
			& Micro-F1 & Macro-F1  & Micro-F1 & Macro-F1&  Micro-F1 & Macro-F1 & Micro-F1 & Macro-F1 \\

			\midrule
			IR-TF-IDF  & 0.53 & 0.52 & 0.49 & 0.48
			& 0.56 & 0.54  &0.65 & 0.58 \\
			Dataless &0.61 &0.53 & 0.50 & 0.47
			&0.59&0.37 &0.71 & 0.48 \\
			Word2Vec &0.33&0.33 & 0.51&0.45
			&0.69&0.47 &0.92&0.83\\
			Doc2Cube&0.23&0.23&0.40&0.35&
			0.67&0.34&0.71&0.38 \\
			WeSTClass&0.49&0.46&0.53&0.43
			&0.50&0.36&0.91&0.84\\
			ConWea&0.65&0.64&0.62&0.57
			&0.91&0.79&0.95&\textbf{0.89}\\
			\textbf{ClassKG~(Ours)}& \textbf{0.78} & \textbf{0.77} & \textbf{0.80} & \textbf{0.75} & \textbf{0.92} & \textbf{0.80} & \textbf{0.96} & 0.83 \\
			
			\bottomrule
			
		\end{tabular}
	}
	\caption{Performance comparison on long text datasets with fine-grained and coarse-grained labels.
	}
	\label{tab:results}
\end{table*}

\subsection{Compared Methods}
We compare our method with a wide range of weakly-supervised text classification methods:
1) \textbf{IR-TF-IDF} evaluates the relevance between documents and labels by aggregated TF-IDF values of  keywords. Documents are assigned labels based on their relevance to labels.
2) \textbf{Dataless}~\cite{Dataless} maps label names and documents to the same space of Wikipedia concepts. Documents are classified by the semantic similarity with each label.
3) \textbf{Word2Vec}~\cite{Word2Vec} first learns the word representations in the corpus and label representations are generated by aggregating the vectors of keywords. Each document is labeled with the most similar label.
4) \textbf{Doc2Cube}~\cite{Doc2Cube} leverages label names as  supervision signals and  performs joint embedding of labels, terms and documents to uncover their semantic similarities.
5) \textbf{BERT count} simply counts keywords to generate pseudo labels for training BERT.
6) \textbf{WeSTClass}~\cite{WeSTClass} generates pseudo documents to train a classifier and bootstraps the model  with self-training.
7) \textbf{LOTClass}~\cite{LOTClass} utilizes only label names to perform classification. They use pre-trained LM to find class-indicative words and generalizes the model via self-training.
8) \textbf{ConWea}~\cite{Conwea} leverages BERT to generate contextualized representations of words, which is further utilized to train the classifier and expand seed words.

\subsection{Experimental Settings}
The training and evaluation are performed on NVIDIA RTX 2080Ti.
In the subgraph annotator, we use a three-layer GIN~\cite{GIN}.
We first train it with our self-supervised task $10^6$ iterations and then finetune it 10 epochs. We set the batch size of self-supervision/finetuning to 50/256.
In classifier training, we set the batch size to 4/8 for long/short texts.
Both the subgraph annotator and the text classifier use AdamW~\cite{AdamW} as optimizer.
Their learning rates are 1e-4 and 2e-6, respectively.
The classifier uses bert-base-uncased for short texts and longformer-base-4096 for long texts.
For keywords extraction, we select top 100 keywords per class in each iteration. The hyperparameter $M$ is set to 4.
The keywords set change threshold $\epsilon$ is set to 0.1. 
Our code has already been released.\footnote{\url{https://github.com/zhanglu-cst/ClassKG}}

\def\hs8{0.1em}
\begin{table}[t] \normalsize  
	\centering 	
	\scalebox{0.9}{
	\setlength{\tabcolsep}{0.6mm}{
		\begin{tabular}{c@{\hspace{\hs8}}|c@{\hspace{\hs8}}|c@{\hspace{\hs8}}|c@{\hspace{\hs8}}|c@{\hspace{\hs8}}}
			\toprule
			Methods  & AG News &  DBPedia & IMDB & Amazon \\
			\midrule
			Dataless & 0.696 & 0.634 & 0.505 & 0.501 \\
			BERT count & 0.752 & 0.722 & 0.677 & 0.654\\
			WeSTClass & 0.823 & 0.811 & 0.774 & 0.753 \\
			LOTClass & 0.864 & 0.911 & 0.865 & 0.916 \\
			\textbf{ClassKG~(Ours)}     & \textbf{0.888} & \textbf{0.980 } & \textbf{0.874} & \textbf{0.926}  \\
			\bottomrule
		\end{tabular}
	}
	}
	\caption{Comparison on short text datasets.
	}
	\label{tab:short_texts}
\end{table}

\subsection{Performance Comparison}
\textbf{Long text datasets}.
The evaluation results are summarized in Table~\ref{tab:results}.
Since the datasets are imbalanced, we use micro-f1 and macro-f1 as evaluation metrics.
As we can see, our method achieves SOTA, and outperforms existing weakly supervised methods in most cases.
On 20Newsgroup, which is a much harder dataset, our method exceeds SOTA for all metrics by a large margin. The gap is 13\% in fine-grained classification and 18\% in coarse-grained classification.
Although the NYT dataset is relatively simple, our model still has advantage on three of the four metrics: achieves over 1\% improvement on 3 metrics, degrades a little only on macro-f1 of coarse-grained, due to the extreme imbalance of categories.

\textbf{Short text datasets}.
Results of short text datasets are shown in Table~\ref{tab:short_texts}.
We follow previous works~\cite{LOTClass} to use accuracy as the metric.
We can see that our method outperforms SOTA on all datasets, especially for DBPedia, the improvement is up to 6.9\%.
With only label names as initial keywords, our method achieve almost 90\% accuracy on all datasets. 

\subsection{Ablation Study}
Here, we check the effects of various components and parameters in our framework. Experiments are conducted on 20News with fine-grained labels.

\subsubsection{Effectiveness of Subgraph Annotator}
\begin{figure}[t]
	\centering
	\includegraphics[width=1.0\columnwidth]{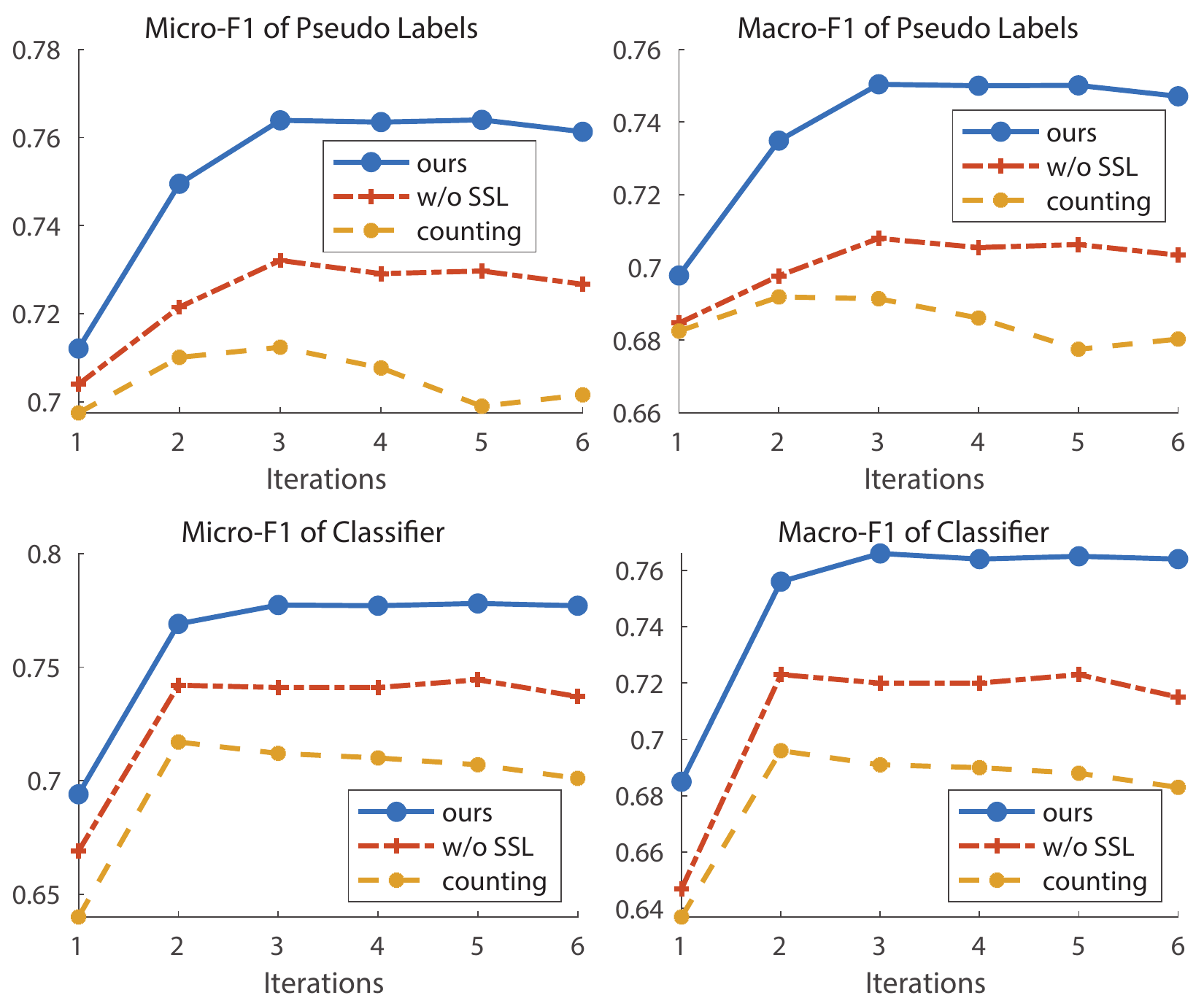} %
	\caption{Comparison with the case without self-supervised learning (SSL) and the case using counting. Results of
	quality of pseudo labels (top) and classification (down) in the first 6 iterations are shown.}
	\label{AB_SA}
\end{figure}

To verify the effectiveness of subgraph annotator $\mathcal{A}$, we compare the results w/wo subgraph annotator $\mathcal{A}$ and w/wo self-supervised learning (SSL).
For the case without $\mathcal{A}$, we use keyword counting to generate pseudo-labels, which is widely used in previous works. 
For the case without self-supervision, we directly finetune the subgraph annotator without self-supervised training.
The results of the first 6 iterations are illustrated in Fig.~\ref{AB_SA}. 

We can see that
1) our method with all components performs much better than the other cases, proving the effectiveness of exploiting the correlation among keywords.
2) For the case using keyword counting, since the correlation among keywords is ignored, the micro/macro-f1 of pseudo labels is the worst, which leads to the worst classification performance.
3) For the case with finetuning but no self-supervised learning, it outperforms keyword counting by 2.5\% and 1.7\% on micro-f1 and macro-f1 of pseudo labels in the 6$^{th}$ iteration, respectively, which further leads to 3.6\% and 3.2\% gain on micro/macro-f1 of classification performance.
4) Our self-supervised learning task can boost performance, exceeding the case without SSL by a large margin of 3.5\% and 4.4\% in terms of micro/macro-f1 of pseudo labels, and 4.0\% and 4.9\% of classification performance.

\subsubsection{Subgraph Annotator Implementation}
We use different GNNs to implement the subgraph annotator, including GCN~\cite{GCN}, GAT~\cite{GAT} and GIN~\cite{GIN}.
For GCN and GAT, we readout the subgraph by averaging all node features in the last layer.
For fair comparison, all GNNs set the layer number to 3.
Performance comparison is given in Table~\ref{tab:AB_GNN}.
We can see that the performance of subgraph annotator is highly related to the selected GNN model, and a more powerful GNN model will lead to higher annotation accuracy.

\def\hs6{0.6em}
\begin{table}[h]\small
	\centering 	
	\scalebox{0.95}{
	\begin{tabular}{c@{\hspace{\hs6}}|c@{\hspace{\hs6}}c@{\hspace{\hs6}}c@{\hspace{\hs6}}c@{\hspace{\hs6}}c@{\hspace{\hs6}}c@{\hspace{\hs6}}}
		\toprule
		Metrics & GNN / Iter & 1 & 2 & 3 & 4 & 5 \\
		\midrule
		\multirow{3}{*}{Micro-F1} &
		GCN   & 0.700 & 0.712 & 0.742 & 0.743 & 0.743     \\
		&GAT  & 0.703 & 0.735 & 0.753 & 0.753 & 0.753   \\
		&GIN  & 0.712 & 0.750 & 0.764 & 0.764 & 0.764  \\
		\hline
		\multirow{3}{*}{Macro-F1} &
		GCN   &  0.684 & 0.697 & 0.726 & 0.726 & 0.726   \\
		&GAT  &  0.689 & 0.719 & 0.737 & 0.737 & 0.737 \\
		&GIN  &  0.698 & 0.735 & 0.750 & 0.750 & 0.750  \\
		
		\bottomrule
	\end{tabular}
}
	\caption{Performance comparison among different subgraph annotator implementations.}
	\label{tab:AB_GNN}
\end{table}

\subsubsection{Effect of the Number of Keywords}
\begin{figure}[t]
	\centering
	\includegraphics[width=1.0\columnwidth]{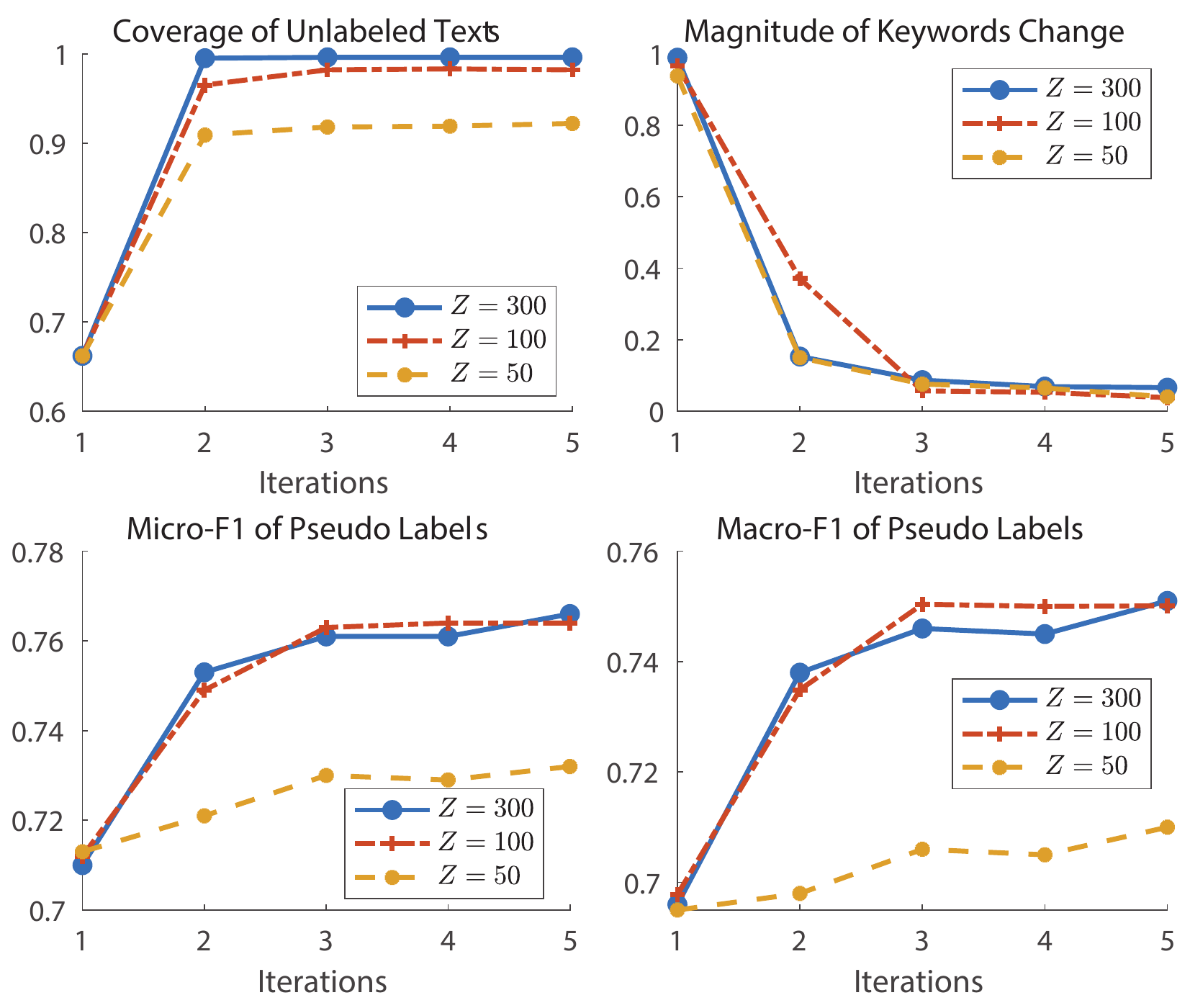} %
	\caption{Effect of keywords number per class $Z$ on the coverage of unlabeled texts, the change of keywords, and the quality of pseudo labels.
	}
	\label{AB_keywords}
\end{figure}

Here, we check the effect of the number of extracted keywords.
We vary the number of extracted keywords per class $Z$ and show the results in Fig.~\ref{AB_keywords}.
We can see that
1) since more keywords will hit more texts, more extracted keywords result in higher text coverage.
2) The change ($\Delta$) of keywords falls below the threshold $\epsilon$ (0.1) in the 3$^{th}$ update for all three keywords number settings. We can assume that  $Z$ has little effect on the number of iterations for model convergence.
3) Increasing the number of keywords from 50 to 100 brings a great performance improvement, while more keywords ($Z=300$) make little change.


\subsubsection{Effect of the IDF Power \textit{M}}
\begin{figure}[h]
	\centering
	\includegraphics[width=1.0\columnwidth]{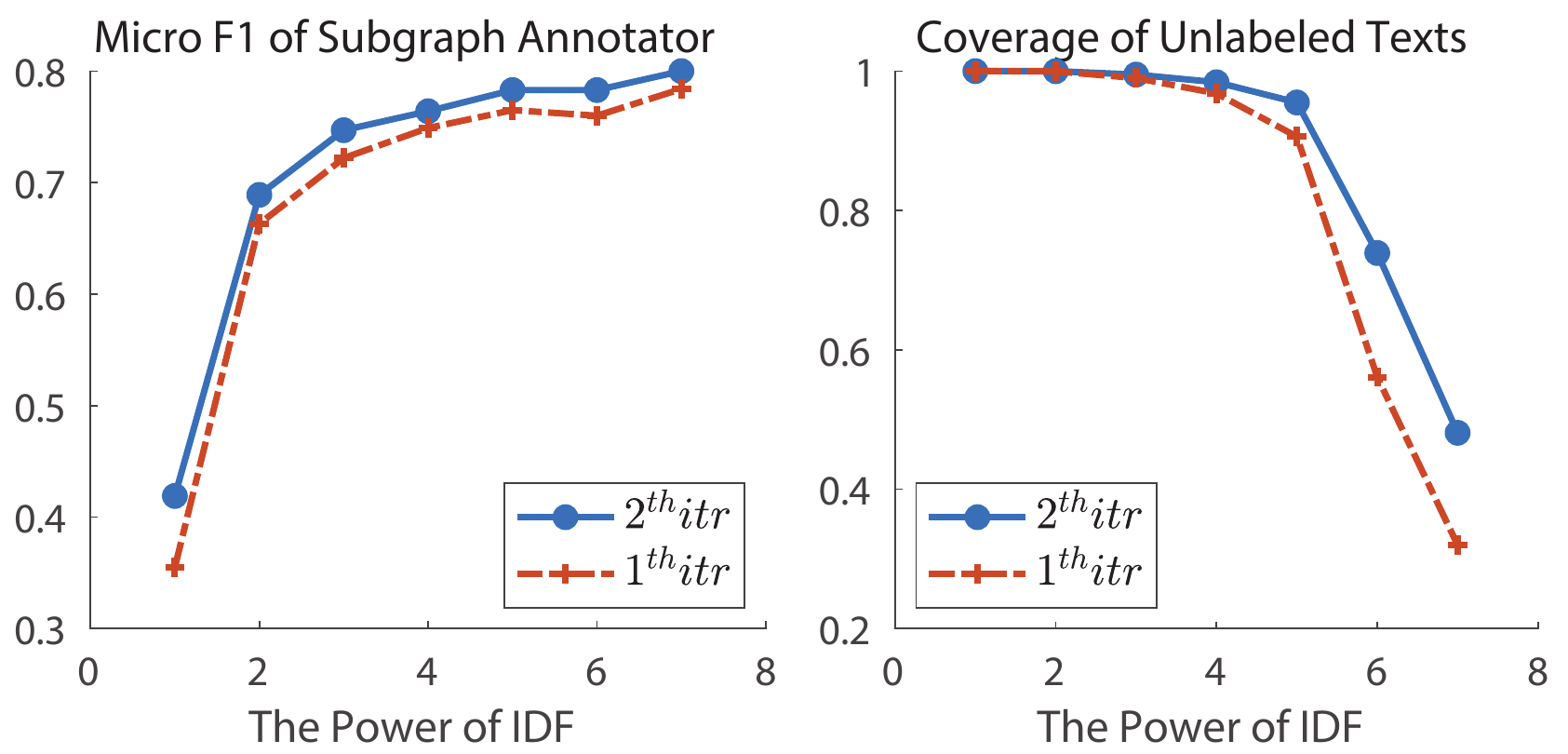} %
	\caption{Effect of the IDF power $M$ on micro-f1 of subgraph annotation and coverage of unlabeled texts.
	}
	\label{AB_IDF}
\end{figure}

We check the effect of the power of IDF $M$ in Eq.~(\ref{eq:word_score}) by changing its value from 1 to 7 for extracting keywords, based on which we train the subgraph annotator and report the micro-f1 of subgraph annotation and the coverage of unlabeled texts.
Results of the $1^{th}$ and $2^{th}$ iterations are shown in Fig.~\ref{AB_IDF}.
With the increase of $M$, the accuracy of labeling also increases, but the coverage decreases.
This is due to that a larger $M$ makes the algorithm extract more uncommon words, thus improving the accuracy while reducing the coverage.

\subsubsection{Effect of Text Classifier Implementation}
Our framework is compatible with any classifier.
Here, we replace the Longformer classifier~\cite{Longformer} for long texts with HAN~\cite{HAN} classifier.
Results are shown in Table~\ref{tab:AB_classifier}.
As we can see, our framework with HAN classifier still achieves good performance, surpassing SOTA by 7\% in micro/macro-f1. 

\def\hs5{0.6em}
\begin{table}[t]\small
	\centering 	
	\begin{tabular}{c@{\hspace{\hs5}}|c@{\hspace{\hs5}}c@{\hspace{\hs5}}c@{\hspace{\hs5}}c@{\hspace{\hs5}}c@{\hspace{\hs5}}c@{\hspace{\hs5}}}
		\toprule
		Metrics & Classifier / Iter & 1 & 2 & 3 & 4 & 5 \\
		\midrule
		\multirow{2}{*}{Micro-F1} &
		HAN         & 0.69 & 0.71  & 0.72 & 0.72 & 0.72     \\
		&Longformer &  0.69 & 0.77 & 0.78 & 0.78 & 0.78  \\
		\hline
		\multirow{2}{*}{Macro-F1} &
		HAN         & 0.68 & 0.70 & 0.70 & 0.71  & 0.71  \\
		&Longformer & 0.68 & 0.76 & 0.77 & 0.76 & 0.77 \\
		
		\bottomrule
	\end{tabular}
	\caption{Results of different classifier implementations.}
	\label{tab:AB_classifier}
\end{table}

\subsubsection{Effects of Hyperparameters}
\begin{figure}[h]
	\centering
	\includegraphics[width=1.0\columnwidth]{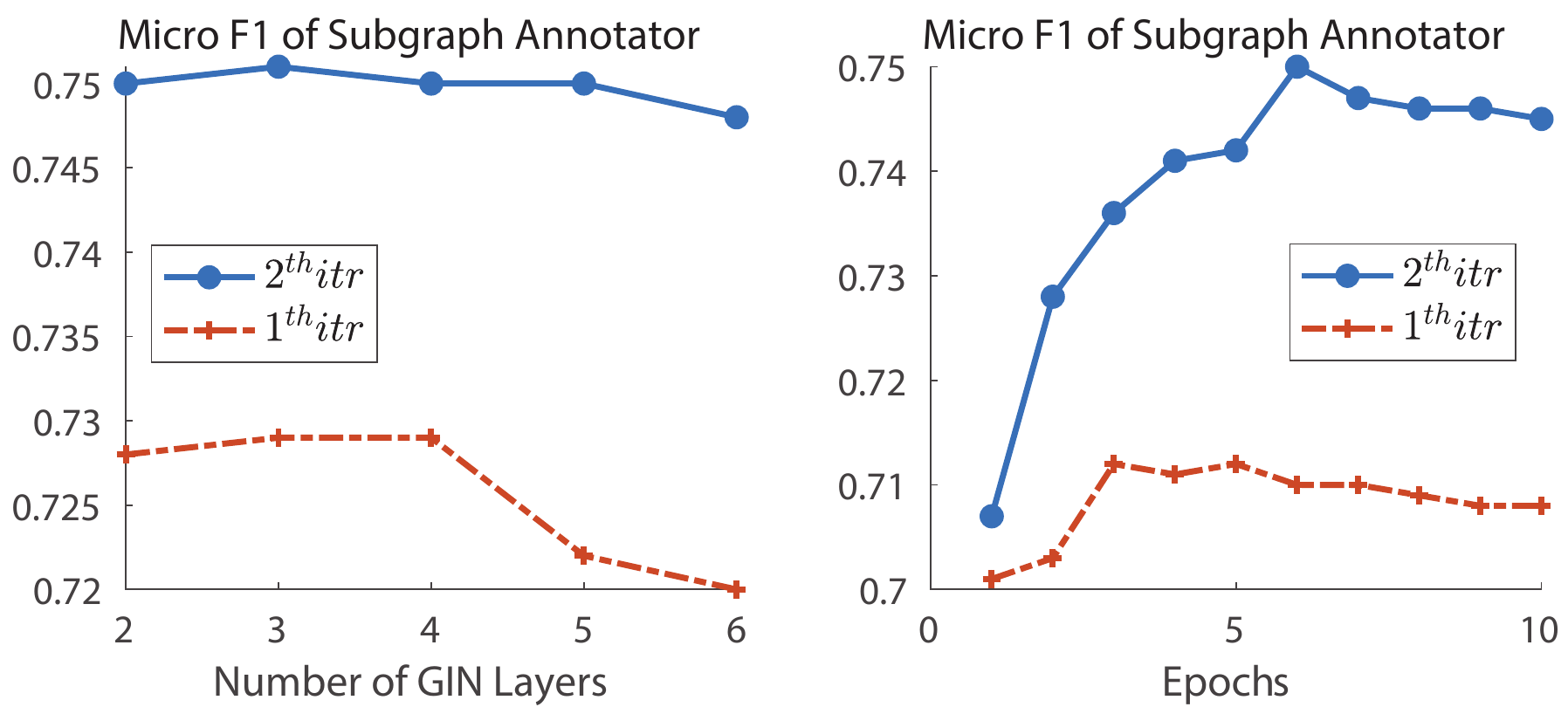} %
	\caption{
		Hyperparameter effect.
	}
	\label{AB_hyper}
\end{figure}

Here, we check the effects of two hyperparameters: the number of GIN layers and the number of epochs for finetuning subgraph annotator. 
The results of the $1^{th}$ and $2^{th}$ iterations are shown in Fig.~\ref{AB_hyper}. We can see that the accuracy of subgraph annotator decreases slightly as the number of GIN layers increases, which may be due to over smoothing.
As finetuning goes, the labeling accuracy decreases slightly, which may be caused by overfitting.

\subsection{Case Study}
Here we present a case study to show the power of our framework.
We take the \textit{technology} class in AG News dataset as an example.
In the beginning, we take ``technology” as the initial keyword.
At the end of the 1st/2nd iteration, the keywords are updated and the top 12 keywords are presented in Table~\ref{tab:case_keywords}.
Obviously, all the 12 keywords extracted by our method are correct, belonging to ``technology'' category.
Furthermore, we check the annotation results in the 2nd iteration.
Some annotations are shown in Table~\ref{tab:case_sentence}. We can see that
our annotator gets more accurate pseudo-labels than keyword counting.

\def\hs9{0.5em}
\begin{table}[h] \small 
	\centering 	
	\begin{tabular}{c@{\hspace{\hs9}}c@{\hspace{\hs9}}}
		\toprule
		Iter     & Top 12 Keywords \\
		\midrule
		0 &   technology\\
		\hline
		1 & \makecell[c]{linux, yahoo, font, mars, nokia, napster\\ peer, adobe, titan, skype, cisco, google }  \\
		\hline
		2 &\makecell[c]{linux, yahoo, font, mars, nokia, adobe \\ titan, skype, abbott, ibm, apple, xbox} \\
		\bottomrule
	\end{tabular}
	\caption{The top 12 keywords of \textit{technology} class.
	}
	\label{tab:case_keywords}
\end{table}

\def\hs9{0.3em}
\begin{table}[h] \footnotesize 
	\centering 	
	\scalebox{0.95}{
	\begin{tabular}{c@{\hspace{\hs9}}|c@{\hspace{\hs9}}c@{\hspace{\hs9}}c@{\hspace{\hs9}}}
		\toprule
		Sentences & Count & $\mathcal{A}$ & GT \\
		\midrule
		\makecell[c]{Google IPO Auction Off \\ to Rocky Start. WASHINGTON ...
			}  &  3 & 2 & 2 \\
		\makecell[c]{Strippers and pole dancers should be\\ banned from performing in stretch...}  & 1 & 0 & 0\\
		\makecell[c]{The Olympics - weapon of mass distraction.\\ As Australians watch their country... }  & 1 & 0 & 0\\
		\bottomrule
	\end{tabular}
	}
	\caption{Some cases that our annotator ($\mathcal{A}$) outperforms keyword counting (Count). `GT': Ground True. `0': politics, `1': sports, `2': business, `3': technology.
	}
	\label{tab:case_sentence}
\end{table}

\section{Conclusion}
In this work, we propose a novel method for weakly-supervised text classification by exploiting the correlation among keywords.
Our method follows an iterative paradigm. In each iteration, we first build a keyword graph and the task of assigning pseudo-labels is transformed into annotating keyword subgraphs.
To accurately annotate subgraphs,
we first train the subgraph annotator with a designed pretext task and then finetune it.
The trained subgraph annotator is used to generate pseudo labels, with which we train a text classifier. Finally, we re-extract keywords from the classification results of the classifier.
Our experiments on both long and short text datasets show that our method outperforms the existing ones.
As for future work, we will focus on improving the proposed method by new mechanisms and network structures.

\textbf{Acknowledgement}.
This work was supported by Alibaba Group through Alibaba Innovative Research Program.

\bibliography{anthology,custom}

\begin{thebibliography}{41}
\expandafter\ifx\csname natexlab\endcsname\relax\def\natexlab#1{#1}\fi

\bibitem[{Agichtein and Gravano(2000)}]{keywords_2000}
Eugene Agichtein and Luis Gravano. 2000.
\newblock Snowball: Extracting relations from large plain-text collections.
\newblock In \emph{Proceedings of the fifth ACM conference on Digital
  libraries}, pages 85--94.

\bibitem[{Badene et~al.(2019)Badene, Thompson, Lorr{\'e}, and
  Asher}]{DP_ACL_Discourse}
Sonia Badene, Kate Thompson, Jean-Pierre Lorr{\'e}, and Nicholas Asher. 2019.
\newblock \href {https://doi.org/10.18653/v1/P19-1061} {Data programming for
  learning discourse structure}.
\newblock In \emph{Proceedings of the 57th Annual Meeting of the Association
  for Computational Linguistics}, pages 640--645, Florence, Italy. Association
  for Computational Linguistics.

\bibitem[{Beltagy et~al.(2020)Beltagy, Peters, and Cohan}]{Longformer}
Iz~Beltagy, Matthew~E. Peters, and Arman Cohan. 2020.
\newblock Longformer: The long-document transformer.
\newblock \emph{arXiv:2004.05150}.

\bibitem[{Chang et~al.(2008)Chang, Ratinov, Roth, and Srikumar}]{Dataless}
Ming-Wei Chang, Lev-Arie Ratinov, Dan Roth, and Vivek Srikumar. 2008.
\newblock Importance of semantic representation: Dataless classification.
\newblock In \emph{Proceedings of the AAAI Conference on Artificial
  Intelligence}, volume~2, pages 830--835.

\bibitem[{Devlin et~al.(2019)Devlin, Chang, Lee, and Toutanova}]{BERT}
Jacob Devlin, Ming-Wei Chang, Kenton Lee, and Kristina Toutanova. 2019.
\newblock \href {https://doi.org/10.18653/v1/N19-1423} {{BERT}: Pre-training of
  deep bidirectional transformers for language understanding}.
\newblock In \emph{Proceedings of the 2019 Conference of the North {A}merican
  Chapter of the Association for Computational Linguistics: Human Language
  Technologies, Volume 1 (Long and Short Papers)}, pages 4171--4186,
  Minneapolis, Minnesota. Association for Computational Linguistics.

\bibitem[{Gabrilovich et~al.(2007)Gabrilovich, Markovitch
  et~al.}]{distant_supervision}
Evgeniy Gabrilovich, Shaul Markovitch, et~al. 2007.
\newblock Computing semantic relatedness using wikipedia-based explicit
  semantic analysis.
\newblock In \emph{International Joint Conference on Artificial Intelligence},
  volume~7, pages 1606--1611.

\bibitem[{He et~al.(2020)He, Fan, Wu, Xie, and Girshick}]{Moco}
Kaiming He, Haoqi Fan, Yuxin Wu, Saining Xie, and Ross Girshick. 2020.
\newblock Momentum contrast for unsupervised visual representation learning.
\newblock In \emph{Proceedings of the IEEE/CVF Conference on Computer Vision
  and Pattern Recognition}, pages 9729--9738.

\bibitem[{Hu et~al.(2020{\natexlab{a}})Hu, Liu, Gomes, Zitnik, Liang, Pande,
  and Leskovec}]{GNN_Pretrain_Strategies}
Weihua Hu, Bowen Liu, Joseph Gomes, Marinka Zitnik, Percy Liang, Vijay Pande,
  and Jure Leskovec. 2020{\natexlab{a}}.
\newblock \href {https://openreview.net/forum?id=HJlWWJSFDH} {Strategies for
  pre-training graph neural networks}.
\newblock In \emph{International Conference on Learning Representations}.

\bibitem[{Hu et~al.(2020{\natexlab{b}})Hu, Dong, Wang, Chang, and
  Sun}]{Gpt-gnn}
Ziniu Hu, Yuxiao Dong, Kuansan Wang, Kai-Wei Chang, and Yizhou Sun.
  2020{\natexlab{b}}.
\newblock Gpt-gnn: Generative pre-training of graph neural networks.
\newblock In \emph{Proceedings of the 26th ACM SIGKDD International Conference
  on Knowledge Discovery \& Data Mining}, pages 1857--1867.

\bibitem[{Jiao et~al.(2020)Jiao, Xiong, Zhang, Zhang, Zhang, and
  Zhu}]{Sub_Graph_Contrast}
Y.~Jiao, Y.~Xiong, J.~Zhang, Y.~Zhang, T.~Zhang, and Y.~Zhu. 2020.
\newblock \href {https://doi.org/10.1109/ICDM50108.2020.00031} {Sub-graph
  contrast for scalable self-supervised graph representation learning}.
\newblock In \emph{2020 IEEE International Conference on Data Mining (ICDM)},
  pages 222--231, Los Alamitos, CA, USA. IEEE Computer Society.

\bibitem[{Kipf and Welling(2017)}]{GCN}
Thomas~N. Kipf and Max Welling. 2017.
\newblock Semi-supervised classification with graph convolutional networks.
\newblock In \emph{International Conference on Learning Representations
  (ICLR)}.

\bibitem[{Kuipers et~al.(2006)Kuipers, Beeson, Modayil, and
  Provost}]{keywords_2006}
Benjamin~J Kuipers, Patrick Beeson, Joseph Modayil, and Jefferson Provost.
  2006.
\newblock Bootstrap learning of foundational representations.
\newblock \emph{Connection Science}, 18(2):145--158.

\bibitem[{Lan et~al.(2020)Lan, Chen, Goodman, Gimpel, Sharma, and
  Soricut}]{ALBERT}
Zhenzhong Lan, Mingda Chen, Sebastian Goodman, Kevin Gimpel, Piyush Sharma, and
  Radu Soricut. 2020.
\newblock \href {https://openreview.net/forum?id=H1eA7AEtvS} {Albert: A lite
  bert for self-supervised learning of language representations}.
\newblock In \emph{International Conference on Learning Representations}.

\bibitem[{Lehmann et~al.(2014)Lehmann, Isele, Jakob, Jentzsch, Kontokostas,
  Mendes, Hellmann, Morsey, Van~Kleef, Auer, and Bizer}]{DBPedia}
Jens Lehmann, Robert Isele, Max Jakob, Anja Jentzsch, Dimitris Kontokostas,
  Pablo Mendes, Sebastian Hellmann, Mohamed Morsey, Patrick Van~Kleef, Sören
  Auer, and Christian Bizer. 2014.
\newblock \href {https://doi.org/10.3233/SW-140134} {Dbpedia - a large-scale,
  multilingual knowledge base extracted from wikipedia}.
\newblock \emph{Semantic Web Journal}, 6.

\bibitem[{Lewis et~al.(2020)Lewis, Liu, Goyal, Ghazvininejad, Mohamed, Levy,
  Stoyanov, and Zettlemoyer}]{BART}
Mike Lewis, Yinhan Liu, Naman Goyal, Marjan Ghazvininejad, Abdelrahman Mohamed,
  Omer Levy, Veselin Stoyanov, and Luke Zettlemoyer. 2020.
\newblock \href {https://doi.org/10.18653/v1/2020.acl-main.703} {{BART}:
  Denoising sequence-to-sequence pre-training for natural language generation,
  translation, and comprehension}.
\newblock In \emph{Proceedings of the 58th Annual Meeting of the Association
  for Computational Linguistics}, pages 7871--7880, Online. Association for
  Computational Linguistics.

\bibitem[{Loshchilov and Hutter(2019)}]{AdamW}
Ilya Loshchilov and Frank Hutter. 2019.
\newblock \href {https://openreview.net/forum?id=Bkg6RiCqY7} {Decoupled weight
  decay regularization}.
\newblock In \emph{International Conference on Learning Representations}.

\bibitem[{Maas et~al.(2011)Maas, Daly, Pham, Huang, Ng, and Potts}]{IMDB}
Andrew~L. Maas, Raymond~E. Daly, Peter~T. Pham, Dan Huang, Andrew~Y. Ng, and
  Christopher Potts. 2011.
\newblock \href {https://www.aclweb.org/anthology/P11-1015} {Learning word
  vectors for sentiment analysis}.
\newblock In \emph{Proceedings of the 49th Annual Meeting of the Association
  for Computational Linguistics: Human Language Technologies}, pages 142--150,
  Portland, Oregon, USA. Association for Computational Linguistics.

\bibitem[{McAuley and Leskovec(2013)}]{Amazon}
Julian McAuley and Jure Leskovec. 2013.
\newblock \href {https://doi.org/10.1145/2507157.2507163} {Hidden factors and
  hidden topics: Understanding rating dimensions with review text}.
\newblock In \emph{Proceedings of the 7th ACM Conference on Recommender
  Systems}, RecSys '13, page 165–172, New York, NY, USA. Association for
  Computing Machinery.

\bibitem[{Mekala and Shang(2020)}]{Conwea}
Dheeraj Mekala and Jingbo Shang. 2020.
\newblock \href {https://doi.org/10.18653/v1/2020.acl-main.30} {Contextualized
  weak supervision for text classification}.
\newblock In \emph{Proceedings of the 58th Annual Meeting of the Association
  for Computational Linguistics}, pages 323--333, Online. Association for
  Computational Linguistics.

\bibitem[{Meng et~al.(2018)Meng, Shen, Zhang, and Han}]{WeSTClass}
Yu~Meng, Jiaming Shen, Chao Zhang, and Jiawei Han. 2018.
\newblock Weakly-supervised neural text classification.
\newblock In \emph{Proceedings of the 27th ACM International Conference on
  Information and Knowledge Management}, pages 983--992.

\bibitem[{Meng et~al.(2019)Meng, Shen, Zhang, and Han}]{AAAI19_hierarchical}
Yu~Meng, Jiaming Shen, Chao Zhang, and Jiawei Han. 2019.
\newblock Weakly-supervised hierarchical text classification.
\newblock In \emph{Proceedings of the AAAI Conference on Artificial
  Intelligence}, volume~33, pages 6826--6833.

\bibitem[{Meng et~al.(2020)Meng, Zhang, Huang, Xiong, Ji, Zhang, and
  Han}]{LOTClass}
Yu~Meng, Yunyi Zhang, Jiaxin Huang, Chenyan Xiong, Heng Ji, Chao Zhang, and
  Jiawei Han. 2020.
\newblock \href {https://doi.org/10.18653/v1/2020.emnlp-main.724} {Text
  classification using label names only: A language model self-training
  approach}.
\newblock In \emph{Proceedings of the 2020 Conference on Empirical Methods in
  Natural Language Processing (EMNLP)}, pages 9006--9017, Online. Association
  for Computational Linguistics.

\bibitem[{Mikolov et~al.(2013)Mikolov, Chen, Corrado, and Dean}]{Word2Vec}
Tomas Mikolov, Kai Chen, Greg Corrado, and Jeffrey Dean. 2013.
\newblock Efficient estimation of word representations in vector space.
\newblock In \emph{International Conference on Learning Representations
  (ICLR)}.

\bibitem[{Noroozi and Favaro(2016)}]{SSL_jigsaw_puzzles}
Mehdi Noroozi and Paolo Favaro. 2016.
\newblock Unsupervised learning of visual representations by solving jigsaw
  puzzles.
\newblock In \emph{European conference on computer vision}, pages 69--84.
  Springer.

\bibitem[{Pappas and Popescu-Belis(2017)}]{HAN}
Nikolaos Pappas and Andrei Popescu-Belis. 2017.
\newblock \href {https://www.aclweb.org/anthology/I17-1102} {Multilingual
  hierarchical attention networks for document classification}.
\newblock In \emph{Proceedings of the Eighth International Joint Conference on
  Natural Language Processing (Volume 1: Long Papers)}, pages 1015--1025,
  Taipei, Taiwan. Asian Federation of Natural Language Processing.

\bibitem[{Pathak et~al.(2016)Pathak, Krahenbuhl, Donahue, Darrell, and
  Efros}]{context_encoders}
Deepak Pathak, Philipp Krahenbuhl, Jeff Donahue, Trevor Darrell, and Alexei~A
  Efros. 2016.
\newblock Context encoders: Feature learning by inpainting.
\newblock In \emph{Proceedings of the IEEE conference on computer vision and
  pattern recognition}, pages 2536--2544.

\bibitem[{Qiu et~al.(2020)Qiu, Chen, Dong, Zhang, Yang, Ding, Wang, and
  Tang}]{GCC}
Jiezhong Qiu, Qibin Chen, Yuxiao Dong, Jing Zhang, Hongxia Yang, Ming Ding,
  Kuansan Wang, and Jie Tang. 2020.
\newblock Gcc: Graph contrastive coding for graph neural network pre-training.
\newblock In \emph{Proceedings of the 26th ACM SIGKDD International Conference
  on Knowledge Discovery \& Data Mining}, pages 1150--1160.

\bibitem[{Ratner et~al.(2016)Ratner, De~Sa, Wu, Selsam, and R{\'e}}]{DP}
Alexander Ratner, Christopher De~Sa, Sen Wu, Daniel Selsam, and Christopher
  R{\'e}. 2016.
\newblock Data programming: Creating large training sets, quickly.
\newblock \emph{Advances in neural information processing systems}, 29:3567.

\bibitem[{Ratner et~al.(2017)Ratner, Bach, Ehrenberg, and Ré}]{Snorkel}
Alexander~J. Ratner, Stephen~H. Bach, Henry~R. Ehrenberg, and Christopher Ré.
  2017.
\newblock \href {https://doi.org/10.1145/3035918.3056442} {Snorkel: Fast
  training set generation for information extraction.}
\newblock In \emph{SIGMOD Conference}, pages 1683--1686.

\bibitem[{Riloff et~al.(2003)Riloff, Wiebe, and Wilson}]{keywords_2003}
Ellen Riloff, Janyce Wiebe, and Theresa Wilson. 2003.
\newblock \href {https://www.aclweb.org/anthology/W03-0404} {Learning
  subjective nouns using extraction pattern bootstrapping}.
\newblock In \emph{Proceedings of the Seventh Conference on Natural Language
  Learning at {HLT}-{NAACL} 2003}, pages 25--32.

\bibitem[{Rong et~al.(2020)Rong, Bian, Xu, Xie, WEI, Huang, and
  Huang}]{Graph_Transformer}
Yu~Rong, Yatao Bian, Tingyang Xu, Weiyang Xie, Ying WEI, Wenbing Huang, and
  Junzhou Huang. 2020.
\newblock \href
  {https://proceedings.neurips.cc/paper/2020/file/94aef38441efa3380a3bed3faf1f9d5d-Paper.pdf}
  {Self-supervised graph transformer on large-scale molecular data}.
\newblock In \emph{Advances in Neural Information Processing Systems},
  volume~33, pages 12559--12571. Curran Associates, Inc.

\bibitem[{Rosenberg et~al.(2005)Rosenberg, Hebert, and
  Schneiderman}]{self-training}
Chuck Rosenberg, Martial Hebert, and Henry Schneiderman. 2005.
\newblock \href {https://doi.org/10.1109/ACVMOT.2005.107} {Semi-supervised
  self-training of object detection models}.
\newblock In \emph{2005 Seventh IEEE Workshops on Applications of Computer
  Vision (WACV/MOTION'05) - Volume 1}, volume~1, pages 29--36.

\bibitem[{Shen et~al.(2021)Shen, Qiu, Meng, Shang, Ren, and Han}]{taxoclass}
Jiaming Shen, Wenda Qiu, Yu~Meng, Jingbo Shang, Xiang Ren, and Jiawei Han.
  2021.
\newblock \href {https://doi.org/10.18653/v1/2021.naacl-main.335}
  {{T}axo{C}lass: Hierarchical multi-label text classification using only class
  names}.
\newblock In \emph{Proceedings of the 2021 Conference of the North American
  Chapter of the Association for Computational Linguistics: Human Language
  Technologies}, pages 4239--4249, Online. Association for Computational
  Linguistics.

\bibitem[{Shu et~al.(2020)Shu, Mukherjee, Zheng, Awadallah, Shokouhi, and
  Dumais}]{email_intent}
Kai Shu, Subhabrata Mukherjee, Guoqing Zheng, Ahmed~Hassan Awadallah, Milad
  Shokouhi, and Susan Dumais. 2020.
\newblock Learning with weak supervision for email intent detection.
\newblock In \emph{Proceedings of the 43rd International ACM SIGIR Conference
  on Research and Development in Information Retrieval}, pages 1051--1060.

\bibitem[{Song and Roth(2014)}]{dataless_hierarchical}
Yangqiu Song and Dan Roth. 2014.
\newblock On dataless hierarchical text classification.
\newblock In \emph{Proceedings of the AAAI Conference on Artificial
  Intelligence}, volume~28.

\bibitem[{Tao et~al.(2015)Tao, Zhang, Chen, Jiang, Hanratty, Kaplan, and
  Han}]{Doc2Cube}
Fangbo Tao, Chao Zhang, Xiusi Chen, Meng Jiang, Tim Hanratty, Lance Kaplan, and
  Jiawei Han. 2015.
\newblock Doc2cube: Automated document allocation to text cube via
  dimension-aware joint embedding.
\newblock \emph{Dimension}, 2016:2017.

\bibitem[{Veličković et~al.(2018)Veličković, Cucurull, Casanova, Romero,
  Liò, and Bengio}]{GAT}
Petar Veličković, Guillem Cucurull, Arantxa Casanova, Adriana Romero, Pietro
  Liò, and Yoshua Bengio. 2018.
\newblock \href {https://openreview.net/forum?id=rJXMpikCZ} {Graph attention
  networks}.
\newblock In \emph{International Conference on Learning Representations}.

\bibitem[{Wang et~al.(2021)Wang, Mekala, and Shang}]{X-class}
Zihan Wang, Dheeraj Mekala, and Jingbo Shang. 2021.
\newblock \href {https://doi.org/10.18653/v1/2021.naacl-main.242} {{X}-class:
  Text classification with extremely weak supervision}.
\newblock In \emph{Proceedings of the 2021 Conference of the North American
  Chapter of the Association for Computational Linguistics: Human Language
  Technologies}, pages 3043--3053, Online. Association for Computational
  Linguistics.

\bibitem[{Xu et~al.(2019)Xu, Hu, Leskovec, and Jegelka}]{GIN}
Keyulu Xu, Weihua Hu, Jure Leskovec, and Stefanie Jegelka. 2019.
\newblock \href {https://openreview.net/forum?id=ryGs6iA5Km} {How powerful are
  graph neural networks?}
\newblock In \emph{International Conference on Learning Representations}.

\bibitem[{Yin et~al.(2019)Yin, Hay, and Roth}]{Zero-shot_Text_Classification}
Wenpeng Yin, Jamaal Hay, and Dan Roth. 2019.
\newblock \href {https://doi.org/10.18653/v1/D19-1404} {Benchmarking zero-shot
  text classification: Datasets, evaluation and entailment approach}.
\newblock In \emph{Proceedings of the 2019 Conference on Empirical Methods in
  Natural Language Processing and the 9th International Joint Conference on
  Natural Language Processing (EMNLP-IJCNLP)}, pages 3914--3923, Hong Kong,
  China. Association for Computational Linguistics.

\bibitem[{Zhang et~al.(2015)Zhang, Zhao, and LeCun}]{AG_News}
Xiang Zhang, Junbo Zhao, and Yann LeCun. 2015.
\newblock \href
  {https://proceedings.neurips.cc/paper/2015/file/250cf8b51c773f3f8dc8b4be867a9a02-Paper.pdf}
  {Character-level convolutional networks for text classification}.
\newblock In \emph{Advances in Neural Information Processing Systems},
  volume~28. Curran Associates, Inc.

\end{thebibliography}
\bibliographystyle{acl_natbib}

\appendix



\end{document}